\documentclass[10pt,twocolumn,letterpaper]{article}

\usepackage[pagenumbers]{cvpr} 
\usepackage{tcolorbox}
\usepackage{float}

\definecolor{cvprblue}{rgb}{0.21,0.49,0.74}
\usepackage[pagebackref,breaklinks,colorlinks,allcolors=cvprblue]{hyperref}

\def\paperID{13803} 
\def\confName{CVPR}
\def\confYear{2026}

\title{How does My Model Fail? Automatic Identification and Interpretation of Physical Plausibility Failure Modes with Matryoshka Transcoders}

\author{Yiming Tang\\
National University of Singapore\\
{\tt\small yiming@nus.edu.sg}
\and
Abhijeet Sinha\\
National University of Singapore\\
{\tt\small abhijeet@nus.edu.sg}
\and
Dianbo Liu\\
National University of Singapore\\
{\tt\small dianbo@nus.edu.sg}
}

\begin{document}
\maketitle
\renewcommand{\thefootnote}{\fnsymbol{footnote}}
\footnotetext[1]{We release our scripts in Github: \url{https://github.com/realyimingtangible/Matryoshka-Transcoders}}
\renewcommand{\thefootnote}{\arabic{footnote}}
\begin{abstract}
Although recent generative models are remarkably capable of producing instruction-following and realistic outputs, they remain prone to notable physical plausibility failures. Though critical in applications, these physical plausibility errors often escape detection by existing evaluation methods. Furthermore, no framework exists for automatically identifying and interpreting specific physical error patterns in natural language, preventing targeted model improvements. We introduce Matryoshka Transcoders, a novel framework for the automatic discovery and interpretation of physical plausibility features in generative models. Our approach extends the Matryoshka representation learning paradigm to transcoder architectures, enabling hierarchical sparse feature learning at multiple granularity levels. By training on intermediate representations from a physical plausibility classifier and leveraging large multimodal models for interpretation, our method identifies diverse physics-related failure modes without manual feature engineering, achieving superior feature relevance and feature accuracy compared to existing approaches. We utilize the discovered visual patterns to establish a benchmark for evaluating physical plausibility in generative models. Our analysis of eight state-of-the-art generative models provides valuable insights into how these models fail to follow physical constraints, paving the way for further model improvements.
\end{abstract}    
\section{Introduction}
\label{sec:introduction}








\begin{figure*}
    \centering
    \includegraphics[width=1\linewidth]{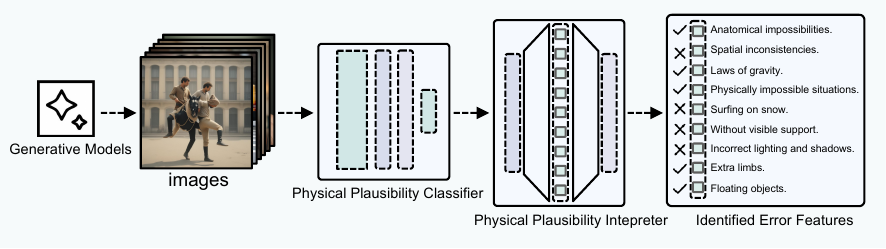}
    \caption{\textbf{Matryoshka Transcoders for Automatic Physical Plausibility Analysis in Generative Models.} We pass images from generative models to a physical plausibility classifier, whose representations are analyzed by Matryoshka Transcoders to extract hierarchical sparse features related to physical plausibility. These features encompasses anatomical impossibilities, spatial inconsistencies, gravity violations, and other physical implausibilities and are interpreted by LMMs automatically.}
    \label{fig:main}
\end{figure*}

Recent advances in generative models have achieved remarkable progress in producing photorealistic and instruction-following images~\citep{rombach2022highresolutionimagesynthesislatent,tian2024visualautoregressivemodelingscalable}. However, these models frequently generate images that appear visually convincing yet violate fundamental physical laws—producing anatomical impossibilities, distorted structures, and malformed text. Such physical plausibility failures significantly undermine the reliability and usability of generative models in high-stakes applications and motivate the development of methods that can identify these failure modes.

Despite their prevalence and importance, physical plausibility errors remain challenging to detect systematically. Existing evaluation metrics primarily focus on semantic alignment between images and text~\citep{radford2021learningtransferablevisualmodels,sampaio2024typescoretextfidelitymetric,yao2024hifi} or aggregate distribution quality~\citep{yu2021frechet,jayasumana2024rethinkingfidbetterevaluation}, leaving specific physical violations largely undetected~\citep{bianchi2024clipmainroadblockfinegrained,schrodi2025effectstriggermodalitygap}. Recent research has shown that even state-of-the-art large multimodal models (LMMs) struggle to consistently identify subtle physical implausibilities when evaluating images holistically~\citep{tang2025humanlikecontentanalysisgenerative}, motivating us to develop methods that can identify such failure modes of generative models. This detection gap leaves such failure modes largely unidentified, hindering the understanding of how some generated images fail in physical plausibility.

Furthermore, no framework exists for automatically identifying and interpreting specific physical error patterns in natural language. Though recent work~\citep{tang2025humanlikecontentanalysisgenerative} has proposed to explore physical plausibility detection by the identification of topic-related visual patterns grounded in natural language, they utilize preliminary interpretability modules and struggle to discover fine-grained error patterns. The absence of interpretable, automatically discovered features means that practitioners lack systematic understanding of the diverse ways generative models violate physical constraints, limiting the development of targeted mitigation strategies.

We introduce Matryoshka Transcoders (See Figure~\ref{fig:main}), a novel framework for the automatic discovery and interpretation of physical plausibility features in generative models. Our approach extends the Matryoshka representation learning paradigm~\citep{bussmann2025learningmultilevelfeaturesmatryoshka} to transcoder architectures~\citep{dunefsky2024transcodersinterpretablellmfeature}, enabling hierarchical sparse feature learning at multiple granularity levels, from coarse-grained violations to fine-grained specific errors. We first assign human annotators to annotate a physical plausibility dataset, including a set of generated images that are marked as physically plausible or not. We use this dataset to train a physical plausibility classifier that contains physical plausibility features, and develop Matryoshka Transcoders on the latent representations of the classifier. The Matryoshka Transcoder identifies specific physical plausibility features, which are then interpreted by LMMs to get their human understandable explanations.

Our experiments demonstrate that Matryoshka Transcoders successfully discover diverse physical plausibility patterns without manual feature engineering, achieving superior feature relevance and accuracy compared to existing methods including standard transcoders, sparse autoencoders, and prior interpretability approaches. We utilize these identified patterns related to physical plausibility to establish the first interpretation-based benchmark for evaluating physical plausibility in generative models. Our analysis of eight state-of-the-art generative models, including SDXL variants~\citep{meng2022sdeditguidedimagesynthesis}, DALL-E3~\citep{dalle3}, FLUX~\citep{FLUX.1-dev}, Kolors~\citep{Kolors-diffusers}, and Stable-Cascade~\citep{stablecascade2025}, reveals systematic failure modes and provides actionable insights into how these models violate physical constraints, from textual distortions to anatomical impossibilities. These findings pave the way for targeted model improvements and more physically plausible generative systems.

Our contributions are listed as below: 

\begin{itemize}
    \item \textbf{Targeted Feature Discovery.} We present the first automated pipeline to identify and interpret visual features in ML models relevant to a human-assigned topic.
    \item \textbf{Matryoshka Transcoders.} We extend the Matryoshka paradigm to transcoder architectures, enabling hierarchical sparse feature learning at multiple granularity levels.
    \item \textbf{Feature Relevance Metrics.} We introduce two metrics, \textit{population-based relevance} and \textit{description-based relevance}, to quantitatively evaluate how well discovered features align with human-assigned topics.
    \item \textbf{Benchmarking Physical Plausibility.} We provide the first interpretation-based benchmark for evaluating existing generative models' capability in producing physically plausible images.
    \item \textbf{Insights on Generative Model Failures.} Our approach discovers diverse physical plausibility features, providing valuable insights into how generative models fail.
\end{itemize}
\section{Related Works}
\label{sec:related}

\subsection{Mechanistic Interpretability}
Mechanistic Interpretability~\citep{bereska2024mechanisticinterpretabilityaisafety,rai2025practicalreviewmechanisticinterpretability} is a research field that aims to reverse-engineer neural networks into human-understandable algorithms and concepts. Modern machine learning models represent diverse information in compact representation spaces where individual neurons activate for multiple unrelated concepts—a phenomenon known as polysemanticity—making it difficult to interpret what each neuron represents~\citep{elhage2022toymodelssuperposition}. A central challenge is decomposing polysemantic neurons within machine learning models into monosemantic neurons that activate for specific interpretable features~\citep{park2024linearrepresentationhypothesisgeometry}. To address this, Sparse Dictionary Learning~\citep{braun2024identifyingfunctionallyimportantfeatures,sharkey2025openproblemsmechanisticinterpretability} has emerged as a powerful approach for automatically discovering interpretable features in neural network activations. Sparse Autoencoders (SAEs)~\citep{cunningham2023sparseautoencodershighlyinterpretable, bricken2023monosemanticity, lieberum2024gemmascopeopensparse} and their variations~\citep{rajamanoharan2024jumpingaheadimprovingreconstruction, gao2024scalingevaluatingsparseautoencoders, bussmann2024batchtopksparseautoencoders} reconstruct neural representations with high sparsity constraints to extract interpretable features. Matryoshka SAEs~\citep{bussmann2025learningmultilevelfeaturesmatryoshka} further improve upon SAEs by simultaneously training nested dictionaries of increasing size, enabling hierarchical feature learning at multiple granularities. Beyond activation reconstruction, Transcoders~\citep{dunefsky2024transcodersinterpretablellmfeature} build sparse mappings between layer activations to identify layer-specific features and have been applied to discover physical plausibility features~\citep{tang2025humanlikecontentanalysisgenerative}.

\subsection{Physical Plausibility in Generative Models}
Although recent generative models are remarkably capable of producing instruction-following and realistic outputs~\citep{rombach2022highresolutionimagesynthesislatent, tian2024visualautoregressivemodelingscalable}, they remain prone to notable physical plausibility failures~\citep{Narasimhaswamy_2024,lu2024handrefinerrefiningmalformedhands}. Existing evaluation approaches primarily focus on semantic alignment or aggregate distribution quality~\citep{radford2021learningtransferablevisualmodels,sampaio2024typescoretextfidelitymetric,yao2024hifi,yu2021frechet,jayasumana2024rethinkingfidbetterevaluation}, leaving specific physical violations largely undetected and unaddressed~\citep{bianchi2024clipmainroadblockfinegrained,kirstain2023pickapicopendatasetuser,zhang2018unreasonableeffectivenessdeepfeatures, schrodi2025effectstriggermodalitygap}. Recently, Language-Grounded Sparse Encoders (LanSE)~\citep{tang2025humanlikecontentanalysisgenerative} proposed to identify interpretable visual patterns for physical plausibility assessment by imitating human cognition~\citep{biederman1987recognition, zeki1999art, nightingale2017can, farid2022creating, Santos03042018, landy2013texture}. However, this approach suffers from limited feature accuracy and struggles to identify specific erroneous patterns. Our work addresses these limitations by leveraging hierarchical transcoder features trained on physical plausibility classifiers, enabling more accurate and fine-grained identification of diverse physical violations.
\section{Method}
\label{sec:method}

\begin{figure*}
    \centering
    \includegraphics[width=0.92\linewidth]{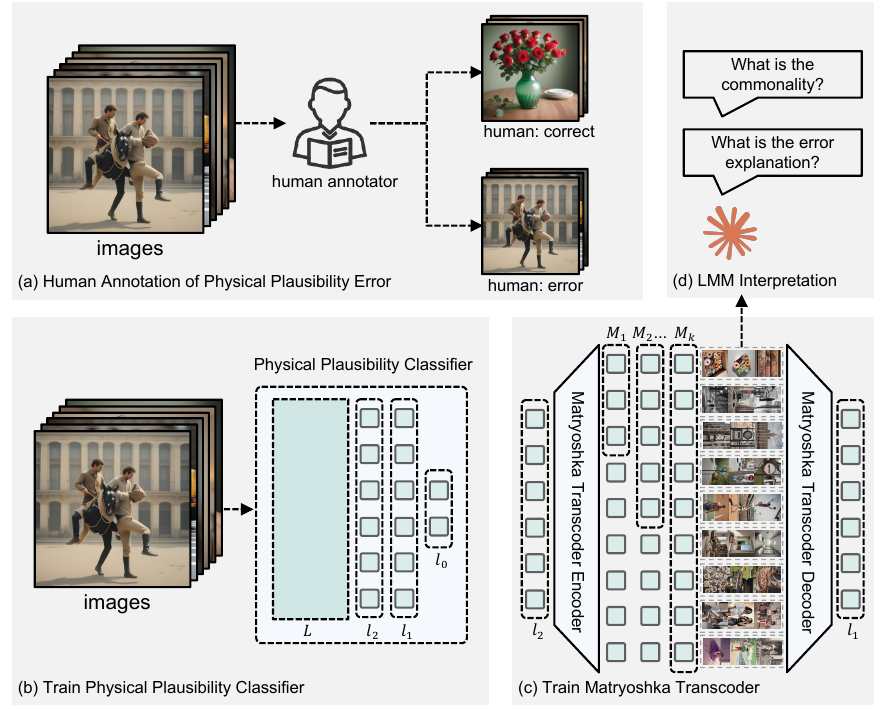}
    \caption{\textbf{Complete pipeline for our proposed method, Matryoshka Transcoders.} Our approach combines four stages: (a) human annotation of correct/error images, (b) training a physical plausibility classifier that possess relevant information, (c) training Matryoshka transcoders to discover sparse, interpretable features at nested granularities from classifier activations, and (d) using large multimodal models to interpret discovered features by identifying visual commonalities and error explanations from maximally activating images.}
    \label{fig:method}
\end{figure*}

\subsection{Overview}
\label{sec:overview}

Our method consists of three stages to automatically identify and interpret physical plausibility features in generated images. First, we train a binary classifier on human-annotated images to distinguish between physically plausible and implausible content, using a frozen CLIP vision encoder with a lightweight classification head (\S\ref{sec:classifier}). Second, we extract intermediate activations from this classifier and train Matryoshka Transcoders to learn hierarchical sparse features that capture physical plausibility-relevant patterns at multiple levels of granularity (\S\ref{sec:matryoshka_transcoder}). Finally, we automatically interpret the learned features using a two-stage prompting strategy with large multimodal models: first identifying the common visual pattern among top-activating images, then analyzing whether this pattern represents a physical plausibility violation (\S\ref{sec:lmm_interpretation}). This pipeline enables scalable discovery and interpretation of diverse physical plausibility errors without manual feature engineering.

\subsection{Physical Plausibility Classifier Training}
\label{sec:classifier}

We employ two independent human annotators to label a dataset of generated images according to their human understanding on the physical plausibility of these images. Each image is assigned a binary label indicating whether it contains physical plausibility errors or is physically plausible. To ensure annotation quality and consistency, we provide annotators with clear guidelines defining physical plausibility violations and include representative examples of common error types. The annotated dataset is augmented with images from natural image datasets, including MSCOCO~\cite{lin2015microsoftcococommonobjects} and Flickr8k~\cite{plummer2016flickr30kentitiescollectingregiontophrase}, which serve as additional negative samples of physically plausible images.

Utilizing this human-annotated dataset, we train a binary classifier to classify natural images as physically plausible or physically implausible. We use CLIP-ViT-Large-patch14~\cite{chen2025contrastivelocalizedlanguageimagepretraining} as the base encoder and attach a two-layer classification head to it. The CLIP vision encoder remains frozen during training to preserve its rich semantic representations learned from large-scale vision-language pretraining. This trained classifier serves as the foundation for extracting intermediate representations that contain physical plausibility information, which will be subsequently analyzed by our Matryoshka Transcoders.

\subsection{Matryoshka Transcoders}
\label{sec:matryoshka_transcoder}

We extract three layers of activations from the trained classifier: base model activations $\mathbf{h}_2 \in \mathbb{R}^{d_2}$, hidden layer activations $\mathbf{h}_1 \in \mathbb{R}^{d_1}$, and classification outputs $\mathbf{h}_0 \in \mathbb{R}^{d_0}$. We train Matryoshka Transcoders on the transformation from $\mathbf{h}_2$ to $\mathbf{h}_1$ to identify physical plausibility-relevant patterns encoded in the hidden representations.

\textbf{Model architecture.} Our Matryoshka Transcoder is a two-layer model with a sparse latent space, consisting of an encoder layer that maps input activations to sparse features:
\begin{equation}
\mathbf{z} = \text{Enc}(\mathbf{h}_2) = \text{ReLU}(W_{\text{enc}} \mathbf{h}_2 + \mathbf{b}_{\text{enc}})
\end{equation}
and a decoder layer that reconstructs the target activations:
\begin{equation}
\bar{\mathbf{h}}_1 = \text{Dec}(\mathbf{z}) = W_{\text{dec}} \mathbf{z} + \mathbf{b}_{\text{dec}}
\end{equation}
where $W_{\text{enc}} \in \mathbb{R}^{d_z \times d_2}$ and $W_{\text{dec}} \in \mathbb{R}^{d_1 \times d_z}$ are the encoder and decoder weight matrices, and $d_z$ is the latent dimension.

\textbf{Training objective.} Our key innovation is to adapt the Matryoshka training paradigm for transcoders. We define a nested sequence of dictionary sizes $\mathcal{M} = \{m_1, m_2, \ldots, m_k\}$ where $m_1 < m_2 < \cdots < m_k = d_z$. For each size $m \in \mathcal{M}$, we enforce sparsity by selecting only the top-$m$ activated latents and compute a reconstruction using these features. The total loss aggregates objectives across all granularity levels:
\begin{equation}
\mathcal{L} = \sum_{m \in \mathcal{M}} \left\|\mathbf{h}_1 - \left( W_{\text{dec}} \mathbf{z}_{0:m} + \mathbf{b}_{\text{dec}}\right)\right\|_2^2
\end{equation}
where $\mathbf{z}_{0:m}$ denotes the first $m$ latent features, and $W_{0:m}^{\text{dec}}$ are the corresponding decoder weights.

\subsection{LMM Interpretation}
\label{sec:lmm_interpretation}

After training the Matryoshka Transcoders, we extract the top-activating images for each learned feature by passing the entire dataset through the trained model. For each feature, we collect the images that pass one activation threshold, along with their associated text captions. To automatically interpret these features in natural language, we employ a two-stage prompting strategy using LMMs.

\textbf{Stage 1: Pattern summarization.} We first identify the common visual pattern captured by each feature. For each feature, we provide the LMM with the image-text pairs and prompt it to summarize the shared visual characteristics. 

\begin{tcolorbox}[title=Pattern Summarization Prompt ($\text{Prompt}_{\text{sum}}$),fonttitle=\bfseries,colback=gray!5!white,colframe=gray!75!black]
You are an expert in visual pattern analysis. Given the following images and their captions: \{image-text pairs\}, analyze the commonalities among them. Concisely summarize one common visual pattern that is present in all instances.
\end{tcolorbox}

\textbf{Stage 2: Physical plausibility analysis.} Given the identified summarization, we prompt the LMM to determine whether this pattern represents a physical plausibility violation and, if so, to describe the violation in natural language.

\begin{tcolorbox}[title=Error Interpretation Prompt ($\text{Prompt}_{\text{interp}}$),fonttitle=\bfseries,colback=gray!5!white,colframe=gray!75!black]
You are an expert in physics and physical reasoning. Here are a list of images and their captions with one common theme, \{pattern from Stage 1\}: \{image-text pairs\} analyze whether these images possess any visual patterns that violate physical plausibility. Consider the following aspects:
\begin{itemize}
\item Anatomic impossibilities (e.g., six fingers in one hand)
\item Distorted structures (e.g., highly distorted and unclear facial expressions)
\end{itemize}
If yes, provide a concise natural language description of such violation possessed by all samples.
\end{tcolorbox}

This two-stage approach enables us to automatically generate interpretable descriptions for each feature learned by the Matryoshka Transcoder.

\section{Experiments}
\label{sec:experiments}
We evaluate Matryoshka Transcoders on their ability to automatically discover interpretable features relevant to physical plausibility and assess their utility for benchmarking generative models. We begin by describing our experimental setup in Section~\ref{sec:setup}, then present qualitative and quantitative analysis of discovered features in Section~\ref{sec:pattern_analysis}, demonstrate the application of our method to benchmark eight generative models in Section~\ref{sec:benchmark}, and conclude with ablation studies validating our design choices in Section~\ref{sec:ablations}.

\subsection{Experimental Setup}
\label{sec:setup}
\textbf{Dataset construction.} We employ human annotators to label generated images as physically plausible or implausible, following clear annotation guidelines for common violation types. Our human-annotated dataset comprises 3,410 images: 888 physically plausible images and 2,522 images containing physical plausibility errors. This dataset is augmented with 5,000 images from MSCOCO~\cite{lin2015microsoftcococommonobjects} as additional negative examples of physically plausible images to train our physical plausibility classifier.

\textbf{Model configuration.} Our classifier uses frozen CLIP-ViT-Large-patch14~\cite{chen2025contrastivelocalizedlanguageimagepretraining} with a two-layer head (768→256→1), trained with AdamW at learning rate $10^{-4}$. Matryoshka Transcoders map CLIP embeddings (768-dim) to hidden activations (256-dim) with nested dictionaries $\mathcal{M} = \{128, 256, 512, 1024, 2048\}$ using top-k sparsities of $\{16, 32, 64, 128, 256\}$ for each granularity.

\textbf{Feature interpretation.} We use Claude Sonnet 4.5~\cite{anthropic2024claude} for automatic feature interpretation via our two-stage prompting strategy (§\ref{sec:lmm_interpretation}), extracting top-20 activating images per feature. The same LMM evaluates feature quality by assessing relevance and accuracy.

\textbf{Computational requirements.} Classifier and Matryoshka Transcoder training each require approximately 1 hour on a single NVIDIA A100 GPU. Feature interpretation via API completes within 20 hours.

\subsection{Physical Plausibility Pattern Identification}
\label{sec:pattern_analysis}

We evaluate the interpretability and relevance of features discovered by Matryoshka Transcoders through both qualitative and quantitative analysis. Our results demonstrate that Matryoshka Transcoders can successfully identifies diverse physical plausibility failure modes and attach highly accurate natural language labels to these failure modes.

\subsubsection{Qualitative results.} 
\begin{figure*}
    \centering
    \includegraphics[width=1\linewidth]{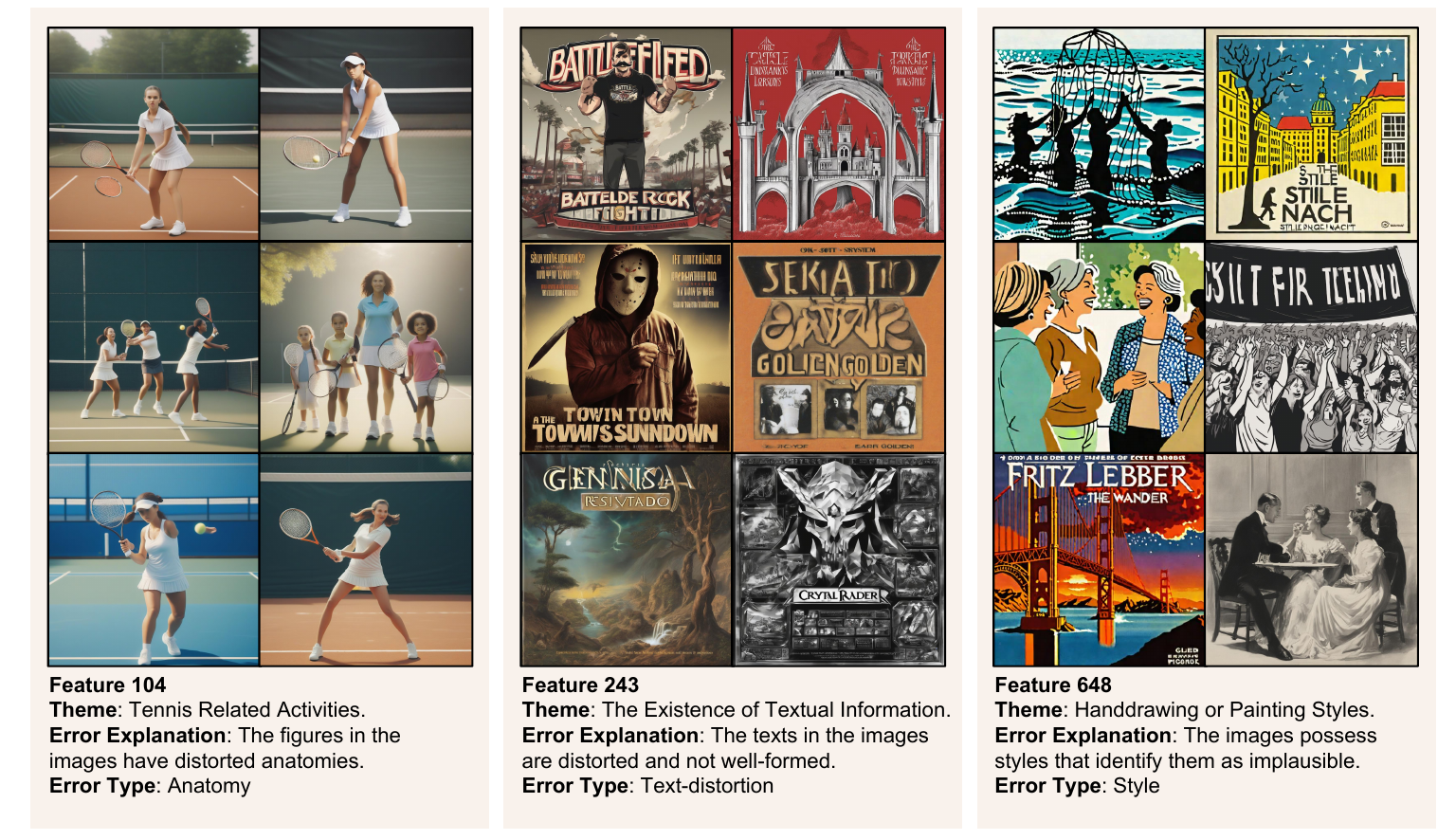}
    \caption{\textbf{Qualitative examples of physical plausibility features discovered by Matryoshka Transcoders.} Each column shows a representative feature with its top-activating images, automatically generated theme, error explanation, and categorized error type.}
    \label{fig:qualitative_examples}
\end{figure*}

Figure~\ref{fig:qualitative_examples} presents representative discovered features with their automatically generated interpretations. Our method identifies diverse violation patterns in varied error types: Feature 104 captures anatomical distortions in tennis scenes, Feature 243 detects malformed text in posters, and Feature 648 recognizes implausible hand-drawn styles. Each feature demonstrates semantic coherence across its activating images, with natural language descriptions characterizing the specific physical plausibility violation.

\subsubsection{Quantitative results.} We evaluate feature quality using two complementary relevance metrics that assess both the selectivity of discovered features and the relevance of their description.

\textbf{Population-based relevance.} For each feature $i$ with activation threshold $\theta$, we compute its topic-relevance score (wrong ratio) as:
\begin{equation}
\label{eq:feature_relevance}
M_i = \frac{|\{x : f_i(x) > \theta \land y(x) = \text{error}\}|}{|\{x : f_i(x) > \theta\}|}
\end{equation}
where $f_i(x)$ is the feature activation on image $x$ and $y(x)$ indicates the error label. This quantifies the fraction of a feature's activating images that contain physical plausibility errors. A feature is classified as topic-relevant if $M_i \geq 0.95$ (see Section~\ref{sec:discussions} for justification), meaning at least 95\% of its activating images exhibit physical violations. The population-based relevance score then measures the concentration of such selective features:
\begin{equation}
\label{eq:population_relevance}
R_{\text{population}} = \frac{|\{i : M_i \geq 0.95\}|}{|\text{Total Features}|}
\end{equation}

\textbf{Description-based relevance.} We also evaluate whether the automatically generated natural language descriptions capture topic-relevant patterns. For each feature, we first generate a description using our two-stage LMM interpretation pipeline (Section~\ref{sec:method}): $\text{description}_i = \text{LMM}(\text{images}_i, \text{Prompt}_{\text{sum}})$. We then assess relevance via:
\begin{equation}
\label{eq:description_relevance}
R_{\text{description}} = \frac{|\{i : \text{LMM}(\text{description}_i) = \text{relevant}\}|}{|\text{Total Features}|}
\end{equation}
where $\text{LMM}(\text{description}_i)$ indicates whether an LMM judges the description as relevant to physical plausibility violations. Together, $R_{\text{population}}$ assesses feature selectivity while $R_{\text{description}}$ evaluates interpretation quality.

We benchmark our method against sparse autoencoders~\citep{makhzani2014ksparseautoencoders}, transcoders~\citep{dunefsky2024transcodersinterpretablellmfeature}, and Matryoshka SAEs~\citep{bussmann2025learningmultilevelfeaturesmatryoshka} as baselines. Specifically, we utilize a large corpus of images generated by different models with diverse captions to identify samples that activate each feature in these interpretability modules, enabling us to compute their relevance scores. Table~\ref{tab:evaluation} shows that Matryoshka Transcoders outperform all baselines on both relevance metrics, achieving 16.00\% population-based and 17.52\% description-based scores. This demonstrates the effectiveness of combining targeted transcoder learning with hierarchical Matryoshka training for discovering physical plausibility features.

\begin{table}[H]
    \centering
    \begin{tabular}{|c|c|c|}
    \hline
    \textbf{Method} &  $R_{\text{population}}$ &  $R_{\text{description}} $ \\
    \hline
    SAE             & 6.54\% & 6.01\% \\
    Matryoshka SAE  & 9.88\% &  5.66\%\\
    Transcoder      & 12.74\% & 11.52\% \\
    \textbf{Matryoshka Transcoder} & \textbf{16.00\%} & \textbf{13.04\%}\\
    \hline
    \end{tabular}
    \caption{\textbf{Main Evaluation Results.} We evaluate methods' capabilities to identify topic-relevant features by comparing their feature relevance scores. Matryoshka Transcoders achieve superior performance on both metrics compared with all other baselines.}
    \label{tab:evaluation}
\end{table}

\begin{figure*}
    \centering
    \includegraphics[width=1\linewidth]{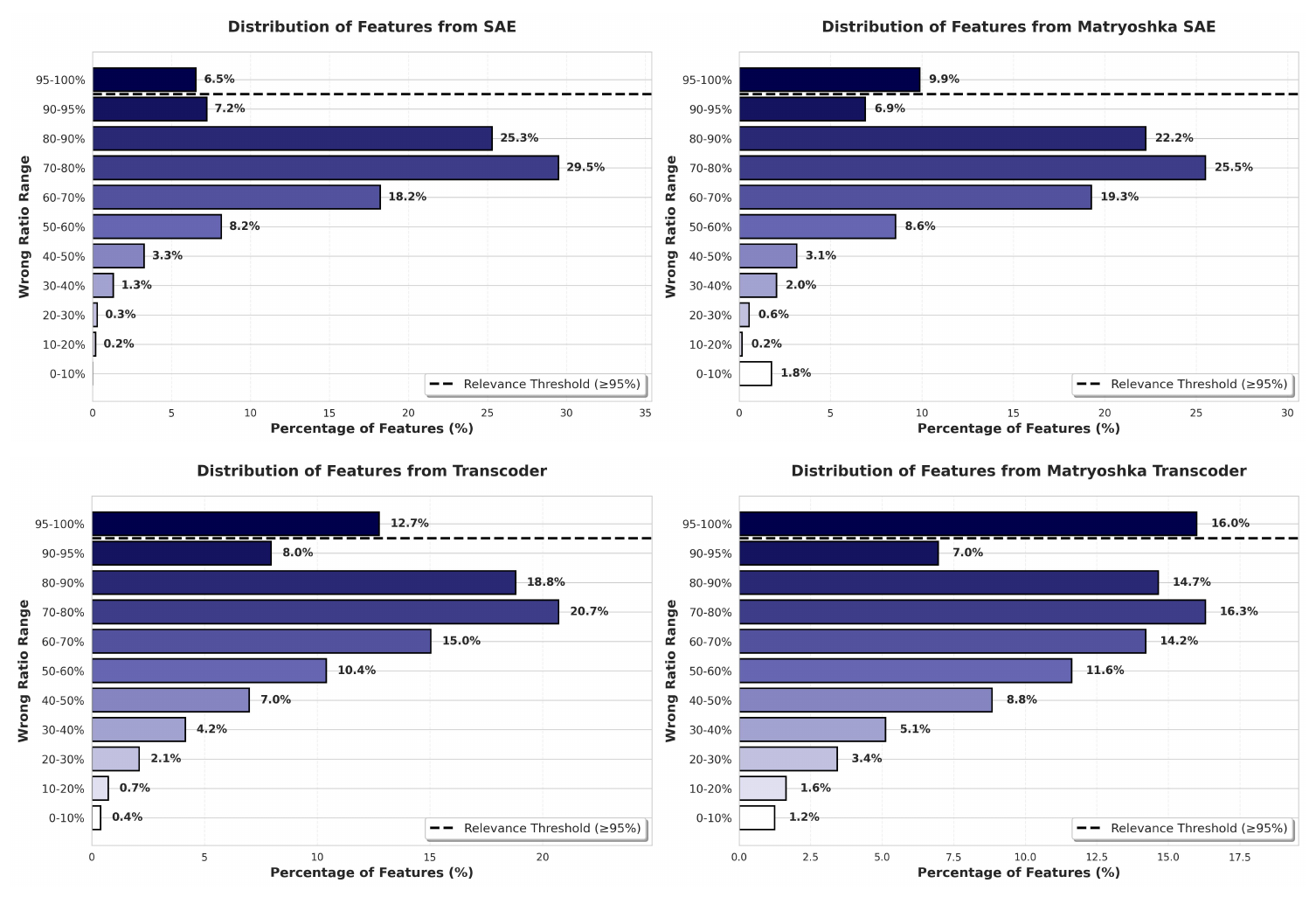}
    \caption{Feature relevance distribution across methods. Percentage distribution of features by their opic-relevance score (wrong ratio) across four sparse dictionary learning methods. Features above the relevance threshold of 0.95 are considered as task-relevant features.}
    \label{fig:feature_distribution}
\end{figure*}

\subsection{Generative Model Benchmark}
\label{sec:benchmark}

We apply discovered physical plausibility features to benchmark eight state-of-the-art generative models. By analyzing which features activate on generated images, we obtain fine-grained diagnostics of each model's physical plausibility failures—revealing not just error frequency but also specific explanations. Table~\ref{tab:benchmark} summarizes our findings.

For each generative model, we compute the mean error feature count as:
\begin{equation}
\text{Error Count} = \frac{1}{N}\sum_{j=1}^{N} |\{i : f_i(x_j) > \theta \land M_i \geq 0.95\}|
\end{equation}
where $N$ is the number of generated images, and $M_i$ is the topic-relevance score. This metric counts how many topic-relevant features activate on each image, averaged across all images. Lower scores indicate better performance, as fewer error-selective features respond to the model's outputs.

\begin{table}[H]
    \centering
    \begin{tabular}{|c|c|}
        \hline
        \textbf{Model Name} & \textbf{Mean Error Count}\\
        \hline
        SDXL-turbo&  74.2 \\
        SDXL-base&  53.1\\
        DALL-E3 & 49.8\\
        SDXL-medium & 48.7 \\
        SDXL-large& 46.8 \\
        Stable-Cascade& 43.1\\
        Kolors& 40.8\\
        \textbf{FLUX}& \textbf{38.0}\\
        \hline
    \end{tabular}
    \caption{\textbf{Physical plausibility evaluation of state-of-the-art generative models.} Error count represents the average number of physical plausibility features activated per generated image across 5,000 test samples. Lower values indicate better performances. FLUX achieves the best physical plausibility, while SDXL-turbo exhibits substantially more violations than other models. Image samples of these generative models can be found in Appendix.}
    \label{tab:benchmark}
\end{table}

\subsection{Ablation Studies}
\label{sec:ablations}

We study how the latent dimension size affects our results.

\begin{figure*}
    \centering
    \includegraphics[width=1\linewidth]{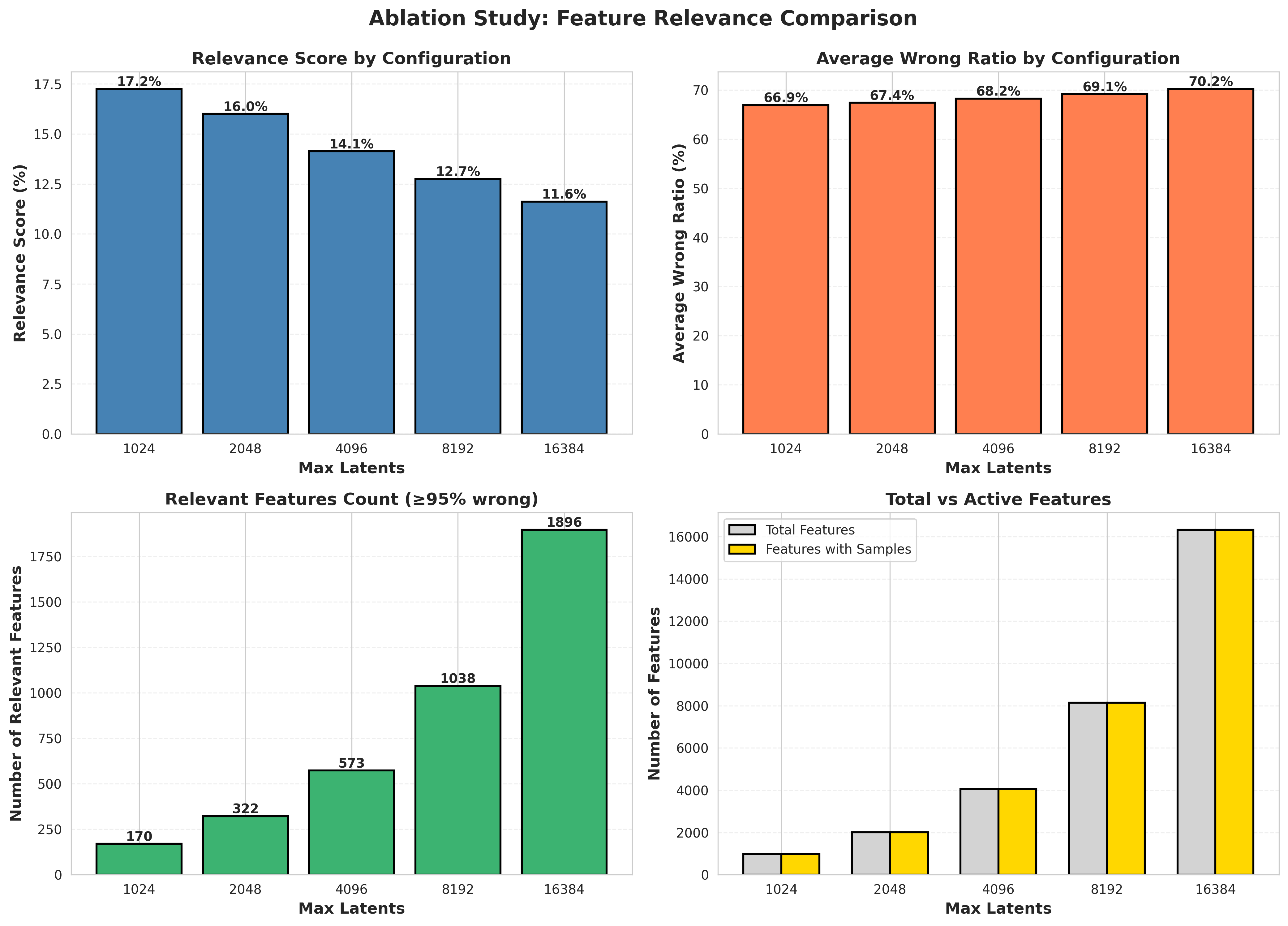}
    \caption{\textbf{Effect of Latent Dimension Size on Feature Discovery.} Ablation study evaluating five dictionary sizes (1024-16384 latents). Smaller dictionaries achieve higher relevance scores (17.2\% vs 11.6\%) but discover fewer total relevant features (170 vs 1,896), revealing a quality-quantity tradeoff. Average wrong ratio remains stable (~67-70\%), indicating consistent feature selectivity across configurations. Nearly all features remain active (94-98\%), showing efficient dictionary utilization regardless of size.}
    \label{fig:ablation_latents}
\end{figure*}

\textbf{Dictionary size.} Figure~\ref{fig:ablation_latents} shows a clear tradeoff: smaller dictionaries (1,024) yield more focused and relevant features (17.2\% relevance), while larger ones (16,384) cover more types of errors but with lower relevance (11.6\%).

\textbf{Feature stability.} The selectivity of error-related features remains stable (about 67–70\%) across all sizes, and most features are active (94–98\%), showing efficient use of the dictionary.

\textbf{In general.} Small dictionaries give concentrated, reliable features; large ones give broader coverage. We use nested sizes (M = {128, 256, 512, 1024, 2048}) to combine both benefits.

\section{Discussion}
\label{sec:discussions}
We introduce Matryoshka transcoders to enable interpretable feature discovery on targeted topics. By training on physical plausibility classifier representations, our method enables automatic discovery of interpretable features relevant to human-assigned topics.

\textbf{The Feature Relevance Bottleneck Phenomenon.} As demonstrated in Figure~\ref{fig:feature_distribution}, feature distributions by the topic-relevance metric $M$ (defined in Equation~\ref{eq:feature_relevance}) reveal a clear bottleneck: features cluster at two extremes ($70\%<M<80\%$ and $M > 95\%$) rather than exhibiting a uniform spread or single-peak distribution. This bottleneck indicates a natural boundary between topic-relevant features that selectively respond to topic-specific visual patterns and task-irrelevant features that merely correlate with the presence of physical plausibility errors. We therefore classify features with $M \geq 0.95$ as topic-relevant. Matryoshka architectures enhance this specialization, yielding 9.9\% and 16.0\% relevant features compared to 6.5\% and 12.7\% for standard architectures. The bottleneck phenomenon justifies focusing interpretability efforts on highly selective features, as intermediate-relevance features likely represent polysemantic patterns that resist coherent interpretation.
\section{Limitations}
\label{sec:limitations}

\begin{itemize}
    \item \textbf{Training integration.} Though our features successfully characterize physical violations, they are not yet incorporated into training pipelines to actively reduce such failures, which lay as one promising future work.
    \item \textbf{Feature coverage.} Certain subtle or rare physical violations may be underrepresented if training data lacks sufficient examples of such patterns, potentially creating evaluation blind spots for edge cases.
    \item \textbf{Feature absorption rate.} We do not quantify the feature absorption phenomenon~\citep{chanin2025absorptionstudyingfeaturesplitting} for our methods. Computing absorption rates for visual inputs is substantially more challenging than for language models due to the high-dimensional, continuous nature of image representations. Future work should develop methods to measure feature completeness in vision models.
    \item \textbf{Evaluation scope.} Our framework focuses on physical plausibility, leaving other important quality dimensions, such as aesthetics, semantic faithfulness, and social biases, for future work.
\end{itemize}
\section{Conclusion}
\label{sec:conclusion}

We introduced Matryoshka Transcoders for automatically discovering and interpreting physical plausibility features in generative models. Our hierarchical approach learns sparse features at multiple granularity levels, enabling interpretable evaluation of physical plausibility and systematically revealing failure modes across generative models. Our work demonstrates that targeted feature discovery is achievable through domain-specific classifiers and developing mechanistic understanding on the classifiers.

{
    \small
    \bibliographystyle{unsrt}
    \bibliography{main}
}

\onecolumn
\newpage
\section*{A. Qualitative Examples of the Generative Model We Evaluate}
\label{appendix A}

\begin{figure*}[h]
    \centering
    \includegraphics[width=1\linewidth]{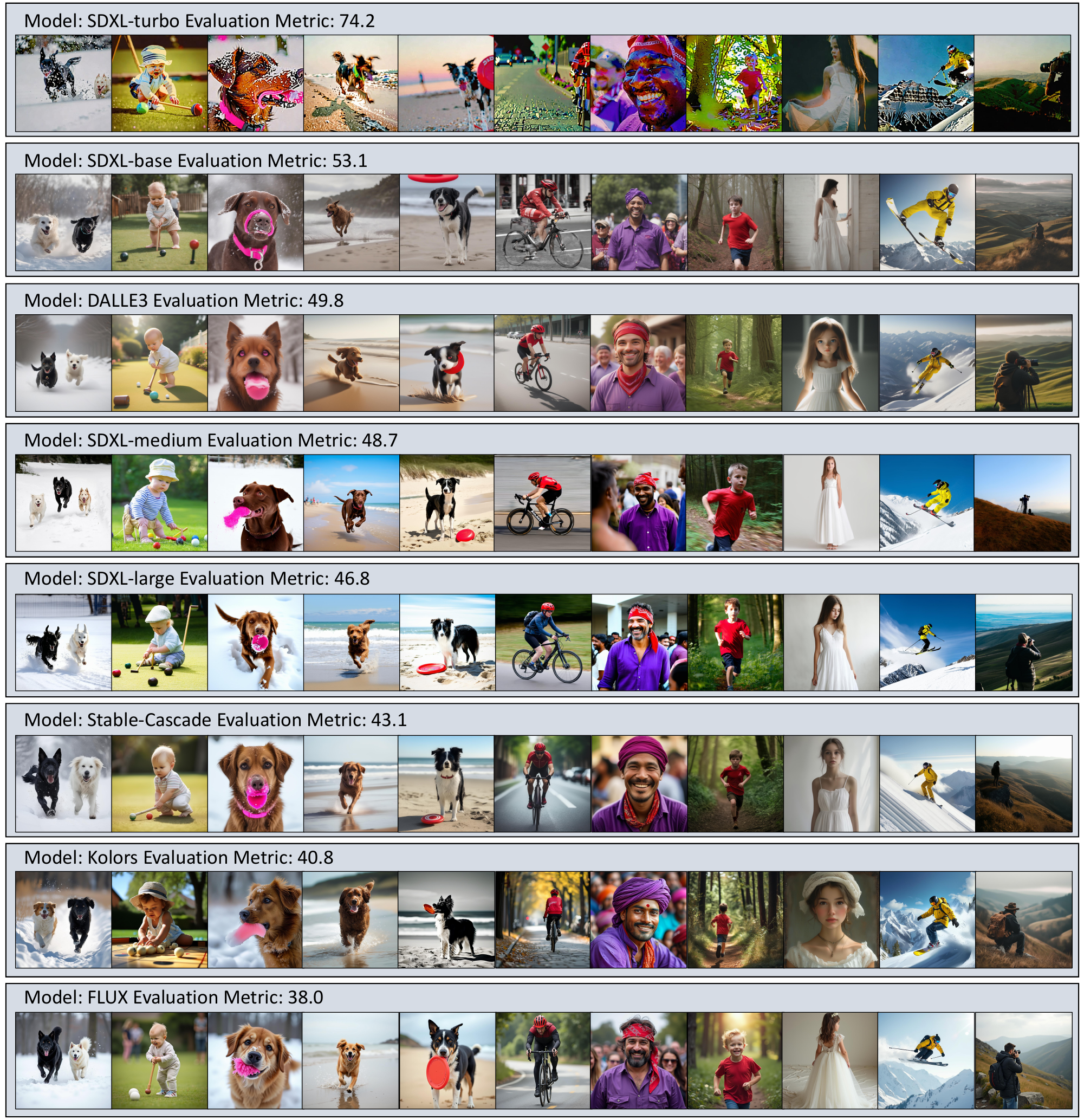}
    \caption{\textbf{Qualitative samples from evaluated generative models ranked by physical plausibility.} Each row shows representative outputs from a different model, ordered by mean error count. Lower error counts correspond to better visual quality and fewer physical violations, with FLUX achieving the best performance (38.0) and SDXL-turbo exhibiting the most plausibility failures (74.2).}
    \label{fig:generative}
\end{figure*}

To complement our quantitative evaluation in Table \ref{tab:benchmark}, we present qualitative samples from all eight evaluated generative models in Figure \ref{fig:generative}.
For each model, we selected a diverse set of representative images spanning common generation scenarios. 
These examples validate that our metric captures meaningful differences in physical plausibility that align with human perception of image quality.

\newpage
\section*{B. Full LMM Prompts}
\label{appendix B}
In this section, we fully describe the prompts we use in our pipelines in detail.

\begin{tcolorbox}[title=Pattern Summarization Prompt ($\text{Prompt}_{\text{sum}}$),fonttitle=\bfseries,colback=gray!5!white,colframe=gray!75!black]
You are an expert in multimodal feature analysis. 

Given the following images and their captions: \{image-text pairs\}.

Analyze the commonalities among these images. Identify: If there exist one common feature that is possessed by all the instances. 

Summarize and output exactly one feature, for example: '[Commonality: Animal wildlife in natural habitats]' and '[Commonality: Strawberry-based dessert or dish]'. 

Only answer in general, do not analyze each image one by one, only generate one single concise phrase. Start answer with '[Commonality:'. End with ']'
\end{tcolorbox}

\begin{tcolorbox}[title=Error Interpretation Prompt ($\text{Prompt}_{\text{interp}}$),fonttitle=\bfseries,colback=gray!5!white,colframe=gray!75!black]

You are an expert in analyzing visual content for physical plausibility errors, anatomical accuracy, and generation artifacts in AI-generated images.

I will show you$ len(samples)$ sample images from a learned feature in an interpretability module trained to detect physical plausibility errors.

**Feature Commonality**: "$commonality$"
**Error Ratio**: ${error_{ratio}}$ of these images contain physical errors ($error_{count}/total_{count}$). 
Be strict with the evaluations. Usually only error ratios larger than 0.95 indicate clear physical plausibility failure mode as these features should not be activated in the presence of any correct images. Other features can be unusable and should be considered as no such error mode.

Here are the sample images:

\{image-text pairs\}.

Based on the $len(samples)$ images shown above and knowing that:
- Feature commonality: "{commonality}"
- ${error_{ratio}}$ contain physical plausibility errors

Analyze whether these images share a **common physical plausibility error or anatomical inaccuracy**. Focus ONLY on:

**Physical Errors:**
- Incorrect number of fingers, toes, or limbs
- Extra or missing body parts
- Distorted anatomy or impossible body proportions
- Unnatural poses or joint configurations
- Incorrect object physics (floating, defying gravity)
- Impossible spatial arrangements or perspectives
- Anatomically incorrect faces or features
- Object inconsistencies or impossible constructions

**Important:**
- Ignore style, artistic choices, or image quality
- Ignore semantic content unless it relates to physical errors
- Focus ONLY on violations of physical or anatomical plausibility

**Response Format:**
If there IS a common physical error across these images:
[Error: Brief description of the specific error]

If there are NO clear physical plausibility errors for all images or if the error ratio does not suggest clear and monosemantic error pattern:
[No common errors]

Your response MUST start with either "[Error:" or "[No common errors]" and end with "]".

\end{tcolorbox}

\newpage
\section*{C. Effect of Dictionary Size on Feature Distribution.}
\label{appendix C}
\begin{figure*}[h]
    \centering
    \includegraphics[width=1\linewidth]{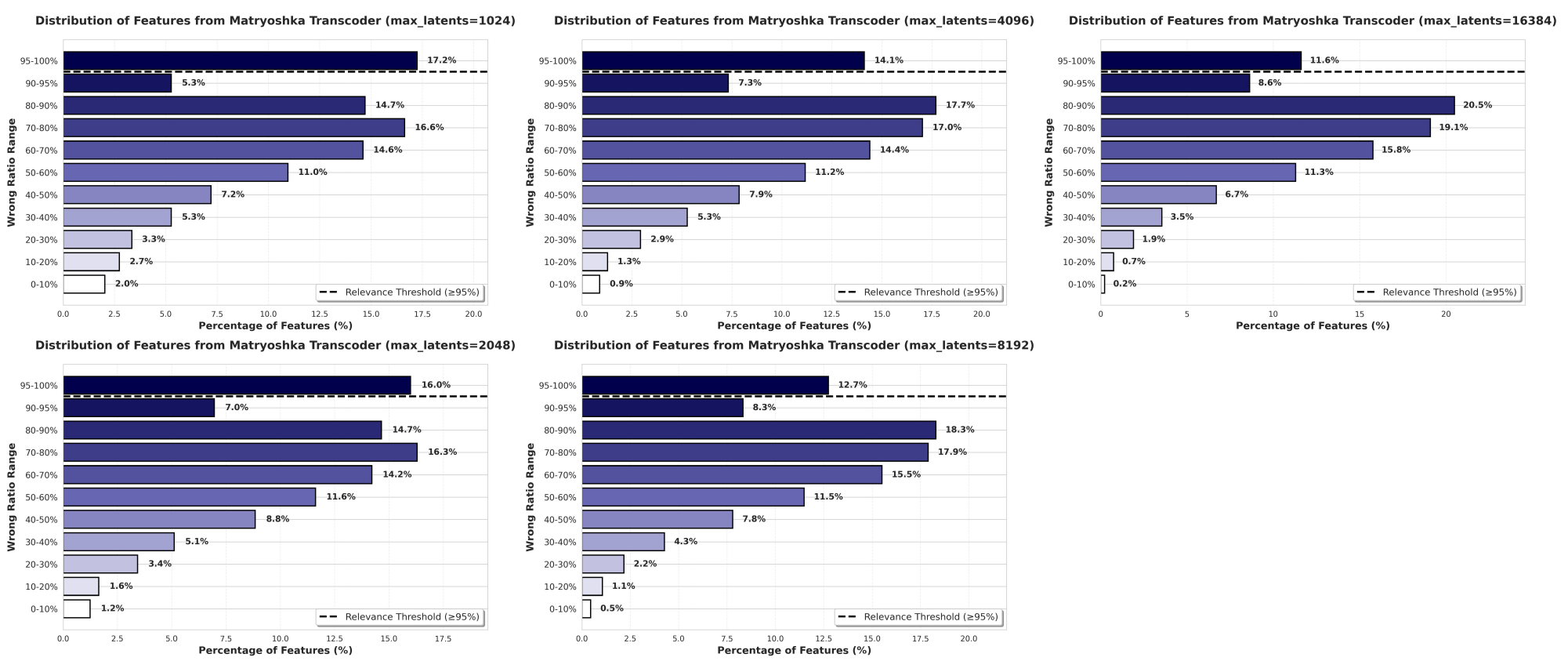}
    \caption{\textbf{Feature relevance distribution across dictionary sizes in Matryoshka Transcoders.} Distribution of features by their topic-relevance score (wrong ratio) for five nested dictionary sizes (1024, 4096, 8192, 16384 latents). Features above the relevance threshold of $0.95$ (indicated by dashed line) are considered topic-relevant. Smaller dictionaries yield higher concentration of relevant features ($17.2\%$ for 1024) but discover fewer total patterns, while larger dictionaries achieve broader coverage (1,896 relevant features for 16384) with lower relevance scores ($11.6\%$). The bimodal distribution consistently shows feature clustering at $70-80\%$ and $\geq95\%$ relevance ranges across all configurations, revealing a natural bottleneck between task-correlated and topic-specific features.}
    \label{fig:dictionary_ablation}
\end{figure*}

Figure~\ref{fig:dictionary_ablation} demonstrates how dictionary size affects the quality and quantity of discovered physical plausibility features. We trained Matryoshka Transcoders with five nested dictionary sizes ranging from 1,024 to 16,384 latents and analyzed the distribution of features by their topic-relevance scores. Smaller dictionaries (1,024 latents) achieve higher feature relevance scores (17.2\%), producing more concentrated and reliable features, but discover fewer total relevant patterns (170 features). In contrast, larger dictionaries (16,384 latents) provide broader coverage of error types, discovering 1,896 relevant features, but with lower average relevance (11.6\%). Notably, all configurations exhibit a consistent bimodal distribution with features clustering at 70-80\% and above 95\% relevance, revealing a natural bottleneck that separates task-correlated features from truly topic-specific features. The average wrong ratio remains stable (67-70\%) across all dictionary sizes, and nearly all features remain active (94-98\%), indicating efficient dictionary utilization. This analysis motivates our nested training approach with $\mathcal{M} = \{128, 256, 512, 1024, 2048\}$, which combines the high relevance of small dictionaries with the comprehensive coverage of larger ones.

\end{document}